# Tracking in Crowd is Challenging: Analyzing Crowd based on Physical Characteristics

Constantinou Miti, Demetriou Zatte, Siraj Sajid Gondal
Relax Tech Sweden

## Introduction:

Currently, the safety of people has become a very important problem [1][2][3] in different places including subway station, universities, colleges, airport, shopping mall and square, city squares. Therefore, considering intelligence event detection systems [4][5][6] is more and urgently required. The event detection method is developed to identify abnormal behavior intelligently [7][8], so public can take action as soon as possible to prevent unwanted activities. The problem is very challenging due to high crowd density in different areas [9][10]. One of these issues is occlusion due to which individual tracking and analysis becomes impossible as shown in Fig. 1. Secondly, more challenging is the proper representation of individual behavior in the crowd. We consider a novel method to deal with these challenges. Considering the challenge of tracking [11][12], we partition complete frame into smaller patches, and extract motion pattern to demonstrate the motion in each individual patch. For this purpose, our work takes into account KLT corners as consolidated features [13][14] to describe moving regions and track these features by considering optical flow method. To embed motion patterns, we develop and consider the distribution of all motion information in a patch as Gaussian distribution, and formulate parameters of Gaussian model as our motion pattern descriptor.

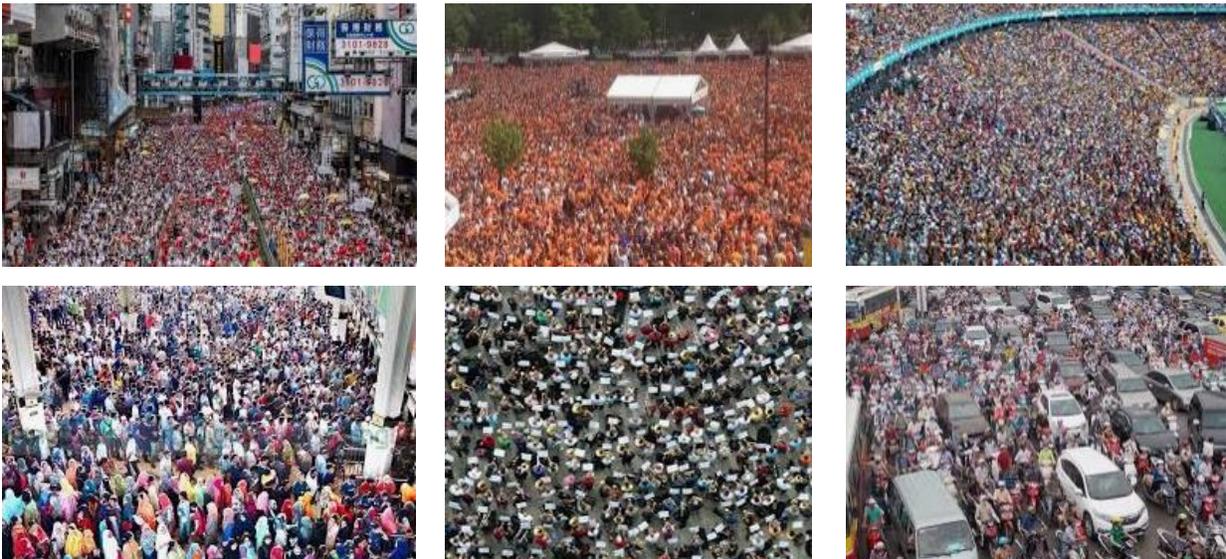

**Fig. 1.** Crowd scenes [1][2][3]. Different crowd areas and scenes are presented where we can see the level of complexities. It is also important to note that the level of occlusion is significantly high.

# Methodology of Our Approach:

There are two methods namely: optical flow [15][16] and spatial-time gradient [17][18]. Optical flow presents motion properly; however, it is driven by huge overheads. Detecting important events in real-time is important for people safety. Therefore, we have to avoid the calculation of optical flow for each pixel. On the other hand calculating spatial-time gradient is not very driven by computational overheads. However, the calculation is based on the extraction of contour features. In crowd areas, object bodies overlap each other and their relative locations change significantly. Therefore, the extraction of contour features is a complex process. Taking into account all these elements, we instead investigate KLT corner as crowd features demonstrate each crowded area. We then extract optical flow features. We also take into account the background subtraction to consolidate our method. In addition to that, we consider the extraction of velocity and direction of each feature produce motion vector as shown in Fig. 2.

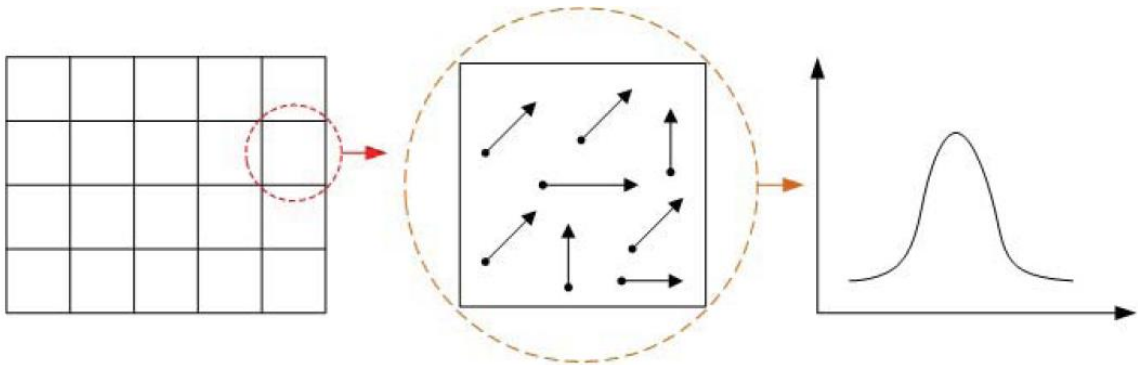

**Fig. 2.** The velocity normalization process shows that each frame is divided into small patches. Moreover, there are motion vectors in the middle and the distribution of motion vector is shown on the right hand side.

To develop our method, we combine coherent motion patterns in a novel way. For example, If we formulate each feature as a pattern model, the amount of features will change significantly. Therefore, it will cost a huge processing overheads. It is important to note that several motion patterns do not match properly in both temporal and spatial order. Therefore, we combine coherent motion patterns as an atomic unit M. In crowd areas, we don't know as a priory the total number or different types of movements. Therefore, we don't know the possible number of coherent regions. Therefore, we formulate the deviation among all motion patterns and develop models, if the smallest deviation is greater than a predefined threshold. Considering this case, we modify the parameters of our method according to the formulation as:

$$M_k = \frac{1}{N_k + 1} P_l + (1 - \frac{1}{N_k + 1}) M_k$$

In the above formulation, $N_k$ is the total number of motion patterns associated with our pattern method $M_k$, and $P_l$ is the updated motion pattern.

Our method does not depend on multiple stages of individual object or person detection or tracking. To keep the computational complexity significantly low, we only consider the foreground areas. In our method, we fuse GMM background subtraction with motion patterns. Therefore, we consider only two holistic features namely: foreground pixel and motion features as illustrated in Fig. 3. We weight our features according to our perspective map to avoid perspective distortions. We also affiliate dense and sparse features to the crowd size. Therefore, we extract the localized corner features and accommodate global corner features. After taking into account the perspective map, the total number of foreground patterns in each video frame is modified according to the formulation as:

$$FeatN_i = \sum_{y=1}^{Y} W_p(y) * N_T(y)$$

In the formulation, $N_T(y)$ is the total number of foreground pixels in the yth row. We introduce local features for global features by considering the relationship between the number of features and the number of foreground pixels. We then develop a weighting model by exploiting the collected features as a crowd events. Our target is to weight the collected features and exploit the calculated features for the magnitudes $d_i$, i = 1…M. M is the total number of frames for the crowd video. We formulate the weight model as:

$$W_d(i) = \frac{d_i - \mu}{\sigma_{max}} + 1$$

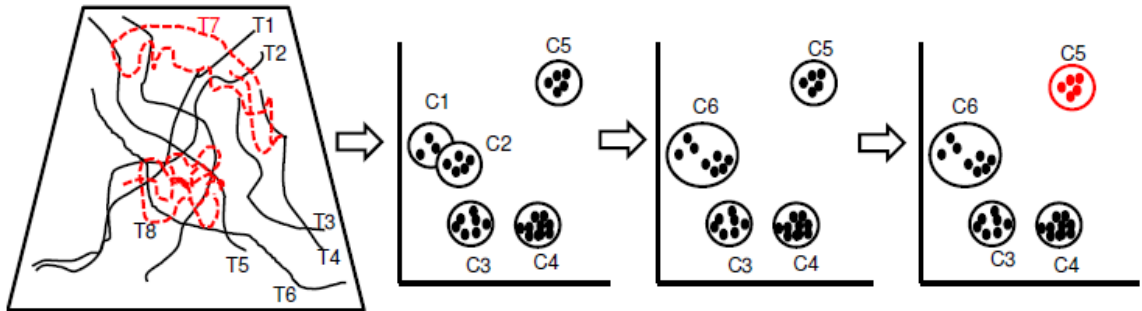

**Fig. 3.** Crowd features integration. Multiple crowd areas are presented on the left hand side. On the right hand side multiple coherent regions are presented.

We developed our method in such a way so that it is not changing according to the changing perspective and changing density of crowded areas. It handles the complexity and non-linearity of our model mapping the collected features to the changing density in the crowd. For the sake of considering more complex situation, we accomodate any unknown errors that could occur in the process of detecting any type of anomaly. We also investigate to exploit Gaussian process regression which adaptively changes and accommodate unknown complexities of crowd when the coherency of crowd changes both locally and globally. Our proposed method carries out crowd analysis for anomaly detection iteratively on smaller time slots thereby efficiently encoding the anomalous crowd situations independent of the complexity of the crowd. Considering the robustness of our method, it can be applied to various applications including environmental monitoring for detecting behaviour anomalies. Our proposed model consisting of multiple stages including: the extraction and combination of features based on crowd motion data, filtering the collected features to be transformed to the next stage, and the detection of anomalous regions.

Our proposed method internally exploits multiple aspects and features including contrast, correlation, energy and homogeneity. In fact, the borders of the video frame, where the remainder of the patches are incomplete, are padded with multiple columns and rows of zeros. A feature model is developed with rows equal to the number of frames and columns equal to the total number of patches. Depending on the area of the person, the patch in which the person is identified is modified with the aforementioned four elements. For a given crowd area, the four important elements are formulated as:

$$\begin{aligned}
\text{Contrast} &= \sum_{i,j} p(i,j)|i-j|^2 \\
\text{Correlation} &= \sum_{i,j} p(i,j)(i-\mu i)(j-\mu j) \\
\text{Energy} &= \sum_{i,j} p(i,j)^2 \\
\text{Homogeneity} &= \sum_{i,j} \frac{p(i,j)}{1+|i-j|}
\end{aligned}$$

## Experimental Analysis and Evaluation:

For experimental evaluation, we use UMN dataset. This dataset consists of three different crowd scenes, and the dataset has 11 videos from these scenes, with a resolution of 240 × 320. Each video sequence represents a normal window slot, for example walking, and an abnormal window slot. The total number of video frames in the scenes 1, scene 2, and scene 3 are 1450, 4415, and 2145, respectively. Initially, we calculate the optical flow between consecutive video frames. We then extract combined features considering different patches. In fact, there are 30x40 patches in a video frame, and the motion patterns are 30x40x16. Subsequently, the collected features from the normal video in the early video frames are collected to build the dictionary.

We used the operating characteristic curve receiver (ROC) to assess the robustness of our method. The ROC is a very good metric to represent the sensitivity and specificity of the collected features. This curve shows the affiliation between sensitivity and septicity through a detailed graph. Huge amount of area under the graph shows better performance. The sensitivity represents the true positive rate (TPR) and the specificity represents the false positive rate (FPR). In fact, the TPR represents the abnormal event that is detected accurately and the FPR represents the normal event that is detected incorrectly.

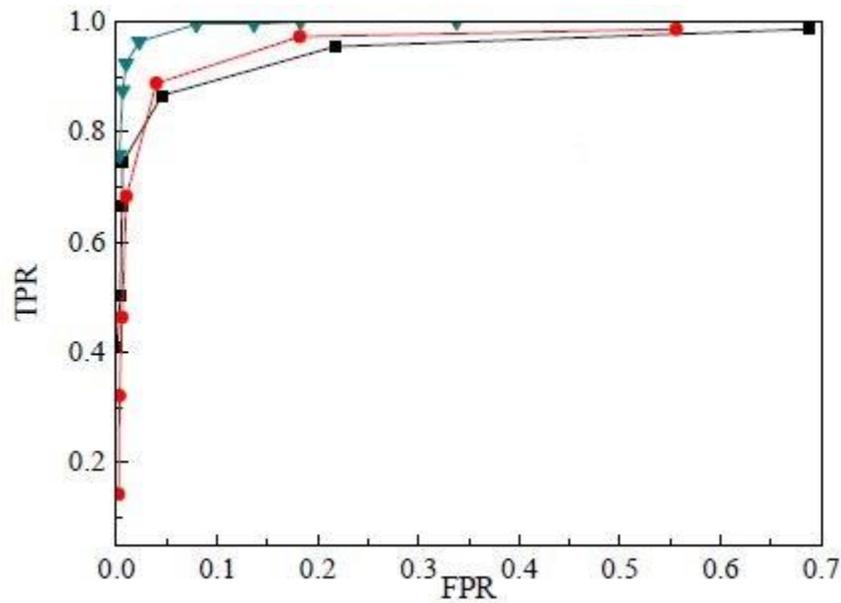

**Fig. 4.** ROC curve for crowd event detection. True positive rate (TPR) and false positive rate (FPR) indicators combined to draw the ROC curve show better performance.

We presented the results in Fig. 4. In this Figure, the result of scene 1 is presented in blue graph, the result of scene 2 is presented in red graph, and the result of scene 3 is presented in black graph. To present the differences of normal and abnormal frames in our method, both TPR and FPR are shown. We show in Fig. 4. That the results are better in term of changing in the graphs. These graphs are stable and smoother. Therefore, our proposed method shows significant improvement.

## Conclusion:

Analysis of abnormal crowd events in different places is significantly important for people safety. However, detecting these behaviors are very challenging due to changing densities and consistencies in crowd movements. In this paper, we have consider several sources of information to propose and develop a robust method for crowd event detection. We integrated the collected features into a single method that results into a method which works very effectively considering a challenging dataset.

## References:

[1] Wang S, Miao Z. Anomaly detection in crowd scene. In IEEE 10th INTERNATIONAL CONFERENCE ON SIGNAL PROCESSING PROCEEDINGS 2010 Oct 24 (pp. 1220-1223). IEEE.

[2] Fradi H, Dugelay JL. Low level crowd analysis using frame-wise normalized feature for people counting. In2012 IEEE International Workshop on Information Forensics and Security (WIFS) 2012 Dec 2 (pp. 246-251). IEEE.

[3] Ullah M, Ullah H, Conci N, De Natale FG. Crowd behavior identification. In2016 IEEE International Conference on Image Processing (ICIP) 2016 Sep 25 (pp. 1195-1199). IEEE.

[4] Rao AS, Gubbi J, Rajasegarar S, Marusic S, Palaniswami M. Detection of anomalous crowd behaviour using hyperspherical clustering. In2014 International Conference on Digital Image Computing: Techniques and Applications (DICTA) 2014 Nov 25 (pp. 1-8). IEEE.

[5] Huang D, Lai JH, Wang CD. Combining multiple clusterings via crowd agreement estimation and multi-granularity link analysis. Neurocomputing. 2015 Dec 25;170:240-50.

[6] Ullah H, Ullah M, Conci N. Real-time anomaly detection in dense crowded scenes. InVideo Surveillance and Transportation Imaging Applications 2014 2014 Mar 5 (Vol. 9026, p. 902608). International Society for Optics and Photonics.

[7] Tripathi G, Singh K, Vishwakarma DK. Convolutional neural networks for crowd behaviour analysis: a survey. The Visual Computer. 2019 May 1;35(5):753-76.

[8] Hao Y, Xu ZJ, Liu Y, Wang J, Fan JL. Effective crowd anomaly detection through spatio-temporal texture analysis. International Journal of Automation and Computing. 2019 Feb 1;16(1):27-39.

[9] Ullah H, Ullah M, Afridi H, Conci N, De Natale FG. Traffic accident detection through a hydrodynamic lens. In2015 IEEE International Conference on Image Processing (ICIP) 2015 Sep 27 (pp. 2470-2474). IEEE.

[10] Kang D, Ma Z, Chan AB. Beyond counting: comparisons of density maps for crowd analysis tasks—counting, detection, and tracking. IEEE Transactions on Circuits and Systems for Video Technology. 2018 May 16;29(5):1408-22.

[11] Kok VJ, Lim MK, Chan CS. Crowd behavior analysis: A review where physics meets biology. Neurocomputing. 2016 Feb 12;177:342-62.

[12] Ullah H, Ullah M, Conci N. Dominant motion analysis in regular and irregular crowd scenes. InInternational Workshop on Human Behavior Understanding 2014 Sep 12 (pp. 62-72). Springer, Cham.

[13] Denis S, Bellekens B, Kaya A, Berkvens R, Weyn M. Large-Scale Crowd Analysis through the Use of Passive Radio Sensing Networks. Sensors. 2020 Jan;20(9):2624.

[14] Ullah M, Ullah H, Cheikh FA. Single shot appearance model (ssam) for multi-target tracking. Electronic Imaging. 2019 Jan 13;2019(7):466-1.

[15] Gkountakos K, Ioannidis K, Tsikrika T, Vrochidis S, Kompatsiaris I. A Crowd Analysis Framework for Detecting Violence Scenes. InProceedings of the 2020 International Conference on Multimedia Retrieval 2020 Jun 8 (pp. 276-280).

[16] Motsch S, Moussaid M, Guillot E, Moreau M, Pettré J, Theraulaz G, Appert-Rolland C, Degond P. Modeling crowd dynamics through coarse-grained data analysis. Mathematical Biosciences and Engineering. 2020 May 23;15(6).

[17] Ullah H. Crowd Motion Analysis: Segmentation, Anomaly Detection, and Behavior Classification (Doctoral dissertation, University of Trento).

[18] Khan SD, Ullah H, Ullah M, Cheikh FA, Beghdadi A. Dimension invariant model for human head detection. In2019 8th European Workshop on Visual Information Processing (EUVIP) 2019 Oct 28 (pp. 99-104). IEEE.

[19] Bilal M, Ullah M, Ullah H. CHEMOMETRIC DATA ANALYSIS WITH AUTOENCODER NEURAL NETWORK. Electronic Imaging. 2019 Jan 13;2019(1):679-1.